\documentclass[11pt, a4paper, logo, twocolumn]{deepmind}

\usepackage{booktabs}
\usepackage{float}
\usepackage{amsfonts}
\usepackage{amsmath}
\usepackage{nicefrac}
\usepackage[capitalize]{cleveref}
\usepackage{xspace}
\usepackage{caption}
\usepackage{subcaption}
\usepackage[authoryear, sort&compress, round]{natbib}
\usepackage[normalem]{ulem}
\usepackage{xcolor}
\usepackage{marginnote}

\newcommand{\grad}{\nabla}
\DeclareMathOperator{\Adv}{\mathrm{Adv}}
\newcommand{\Goals}{\mathcal{G}}
\newcommand{\llc}{\pi_{\mathrm{LLC}}}

\title{Hierarchical Reinforcement Learning in Complex 3D Environments}

\correspondingauthor{bavilapires@deepmind.com}

\keywords{Hierarchical Reinforcement Learning, Partially Observable Markov Decision Processes, Deep Reinforcement Learning}

\renewcommand{\today}{}

\author[*, 1]{Bernardo Avila Pires}
\author[*, 1]{Feryal Behbahani}
\author[*, 1]{Huber Soyer}
\author[*, 1]{Kyriacos Nikiforou}
\author[*, 1]{Thomas Keck}
\author[*, 1]{Satinder Singh}

\affil[*]{Equal contributions}
\affil[1]{DeepMind}

\begin{abstract}

Hierarchical Reinforcement Learning (HRL) agents have the potential to demonstrate appealing capabilities such as planning and exploration with abstraction, transfer, and skill reuse.
Recent successes with HRL across different domains provide evidence that practical, effective HRL agents are possible, even if existing agents do not yet fully realize the potential of HRL.
Despite these successes, visually complex partially observable 3D environments remained a challenge for HRL agents.
We address this issue with Hierarchical Hybrid Offline-Online (H2O2), a hierarchical deep reinforcement learning  agent that discovers and learns to use options from scratch using its own experience.
We show that H2O2 is competitive with a strong non-hierarchical Muesli baseline in the DeepMind Hard Eight tasks and
we shed new light on the problem of learning hierarchical agents in complex environments.
Our empirical study of H2O2 reveals previously unnoticed practical challenges and brings new perspective to the current understanding of hierarchical agents in complex domains.

\end{abstract}

\begin{document}

\maketitle

\section{Introduction}
\label{sec:introduction}

Hierarchical Reinforcement Learning \citep[HRL;][]{barto2003recent,sutton2018reinforcement,pateria2021hierarchical,hutsebaut2022hierarchical} is a framework that could provide us with general and reusable agent representations and behaviors that can exhibit improved exploration and temporal abstraction \citep{nachum2019does}. 
The inspiration comes from humans' ability to break down novel tasks into a sequence of simpler sub-tasks they know how to solve \citep{solway2014optimal}.
This hierarchical approach enables us to transfer our knowledge and reuse our skills to solve new problems.

\paragraph{Contributions.}
In this work we introduce Hierarchical Hybrid Offline-Online (H2O2), a hierarchical deep reinforcement learning agent that discovers and learns to use options from scratch using its own experience.
We show that H2O2 is competitive with a strong non-hierarchical Muesli baseline \citep{hessel2021muesli} in the Hard Eight task suite \cite{gulcehre2019making,worlds2020unity}.
These are challenging sparse-reward tasks in a complex partially observable, first-person 3D environment.
H2O2 employs a combination of  primitive actions and temporally-extended options selected from a continuous option space\footnote{We also provide \href{https://youtube.com/playlist?list=PLlHafZmkZCGmIYSHO9aot75l07kbiuLN0}{videos of the agent} (see \cref{sec:agent-videos} for details)}.
To the best of our knowledge, this is the first hierarchical agent that can be competitive with a strong flat baseline in tasks as complex as the Hard Eight suite, while demonstrably using options to solve tasks.

Our work also sheds new light on the problem of learning hierarchical agents and learning options in complex environments.
We use H2O2 to test a number of hypotheses about its learning and performance in response to changes in its hierarchical design,
and our results reveal previously undetected practical challenges.
While some of our experiments support conclusions in line with the conventional understanding of HRL, others challenged our understanding of hierarchical agents.
For instance, we observed that seemingly beneficial actions such as increasing the agent's option space or allowing it to learn longer options, can actually hurt its performance.

\section{Background and Related Work}
\label{sec:background}

A common HRL approach is to add options to the MDP, turning it into a semi-MDP  \citep[\textbf{SMDP};][]{sutton1999between}, and then use a general-purpose RL algorithm to solve the SMDP \citep{dayan1992feudal,barto2003recent}.
\emph{SMDP agents} decompose into a low-level controller (\textbf{LLC}, which executes the options in the original MDP) and a high-level controller (\textbf{HLC}, which learns to solve the SMDP).
The options may include the actions of the MDP (the \emph{primitive actions}) in addition to temporally extended behaviors \emph{per se}, so that no generality is lost when solving the SMDP instead of solving the MDP ``directly''.

The SMDP strategy effectively changes the problem for the general-purpose RL algorithm.
It is possible to add various capabilities to the general-purpose RL algorithm by augmenting the SMDP with expressive, diverse options \citep{barreto2019option}, additional/alternative state representations \citep{dayan1992feudal,shah2021value}, and even actions to exert fine-grained control over the options \citep{precup2000temporal,barto2003recent}.

Options can be learned from experience,
which can be generated by the hierarchical agent itself \citep[``learned from scratch'';][]{bacon2017option,eysenbach2018diversity,harutyunyan2019termination,wulfmeier2021data,ajay2020opal,hafner2022deep}, or by another agent \citep[for example, an expert;][]{merel2018neural,lynch2020learning,lynch2020grounding}.
The agent can be learned as one unit \citep{bacon2017option,merel2018neural,wulfmeier2021data,ajay2020opal}, but one can also decouple the LLC's option-learning and the HLC's general-purpose RL algorithm \citep{dayan1992feudal,vezhnevets2017feudal,hafner2022deep}.

There have been recent successes in learning \emph{goal-conditioned} options \citep{machado2016learning,lynch2020learning,lynch2020grounding,ajay2020opal,khazatsky2021can, mendonca2021discovering}.
These behaviors are trained to produce specific outcomes (observations or states) in the environment,
and they are commonly learned in hindsight from trajectories of agent \citep{andrychowicz2017hindsight}.
The idea is to identify goals achieved in each trajectory, and use trajectories as demonstrations of behavior that achieves the goal.
The policy can be trained, for example, using behavior cloning \citep[BC;][]{pomerleau1989alvinn} or offline RL \citep{lange2012batch,nachum2018data,fujimoto2019offpolicy,levine2020offline,fu2020d4rl,gulcehre2020rlunplugged}.
The effectiveness of the learned options is largely affected by the choice of policy-learning algorithm, the data available, and how goals are discovered.
For example, BC has been shown to yield effective goal-conditioned policies when used on data generated by experts \citep{lynch2020learning}, but not on non-expert data \citep{ajay2020opal}, whereas offline RL has shown more promise in the latter case.

Discovering both which sub-behaviors to learn and how to combine them can be tackled by pre-learning skills/behaviors with a variety of signals such as expert demonstrations \citep{gupta2020relay}, pseudo-rewards \citep{barreto2019option}, state-space coverage \citep{eysenbach2018diversity,lee2019efficient,islam2019marginalized,pong2020skew}, empowerment \citep{gregor2017variational}, among many others.
Alternatively, the agent can learn its sub-behaviors from its own data, that is, ``from scratch'' \citep{wulfmeier2021data,hafner2022deep}.
This approach has the appeal of being end-to-end, and is philosophically aligned with mainstream deep RL, where agents learn on data that is relevant to the task they are expected to perform \citep{mnih2015human,silver2018general,gulcehre2020rlunplugged,fu2020d4rl}.
It is justified on the grounds that a learning agent will ultimately have to collect novel experience and learn novel sub-behaviors on that experience.

The set of options added to the SMDP can also vary.
If there are only a few options, they can be learnt as separate entities \citep{bacon2017option,wulfmeier2021data}.
A much larger set of options (and, more specifically, goal conditioned policies), on the other hand, can be learned implicitly by encoding the options (goals) in latent space, and treating any element of the latent space as a valid option (goal) \citep{merel2018neural,lynch2020learning,lynch2020grounding,ajay2020opal,hafner2022deep}.
In this case the whole latent space is part of the action space for the SMDP, and the HLC needs to learn to select elements of this latent space.
The complexity of the set of options can be, to a certain extent, limited, by regularizing or constraining the latent output of the option encoders \citep{merel2018neural,lynch2020learning,lynch2020grounding,ajay2020opal,hafner2022deep}.
We are not aware of any successful HRL approaches that encode goals in latent space but do not constrain the latent output of the option encoders in some way.
This suggests that some manner of latent space regularization is essential for deep RL SMDP agents, and this hypothesis is consistent with the empirical findings we present in this work.
We will see in our experiments that not constraining the latent output of the encoder is detrimental H2O2's performance.

We are primarily interested in partially observable environments,
and we adopt the typical deep RL setup for this type of domain:
The agent, at each timestep $t$, must select the action $a_t$ according to a stochastic policy that can depend on the \emph{history} (the sequence of past observations $o_1, \ldots, o_t$ and actions $a_1, \ldots, a_{t-1}$).
This reduces the POMDP to an MDP whose states are the histories of past observations and actions \citep{cassandra1994acting}.
This design choice burdens the deep RL agent with learning to represent histories effectively, but it allows us to use general-purpose deep RL algorithms on both MDPs and POMDPs.
For SMDP agents, both the MDP and the SMDP are treated as above: The LLC acts in a partially observable environment reduced to an MDP over histories, and the HLC acts in a partially observable environment reduced to an SMDP over histories.

\section{Agent Design}
\label{sec:the_agent}

H2O2 is an SMDP agent with options learned from scratch, in offline fashion, that is  \emph{decoupled} from the HLC.
The options are goal-conditioned: The goals are selected in hindsight from experience and encoded into a latent space; then we use offline RL to train an LLC to attain these goals.
The general-purpose deep RL algorithm used for the HLC is Muesli \citep{hessel2021muesli}, which has state-of-the-art performance in the Atari benchmark \citep{bellemare2013arcade}. 
We train the HLC online, through interaction with the environment, as usual for deep RL agents.
Due to our agent's hierarchical design and how its components are trained, we call it \textbf{H}ierarchical \textbf{H}ybrid
\textbf{O}ffline-\textbf{O}nline (\textbf{H2O2}).

\begin{figure}[htb!]
    \centering
    \includegraphics[width=\columnwidth]{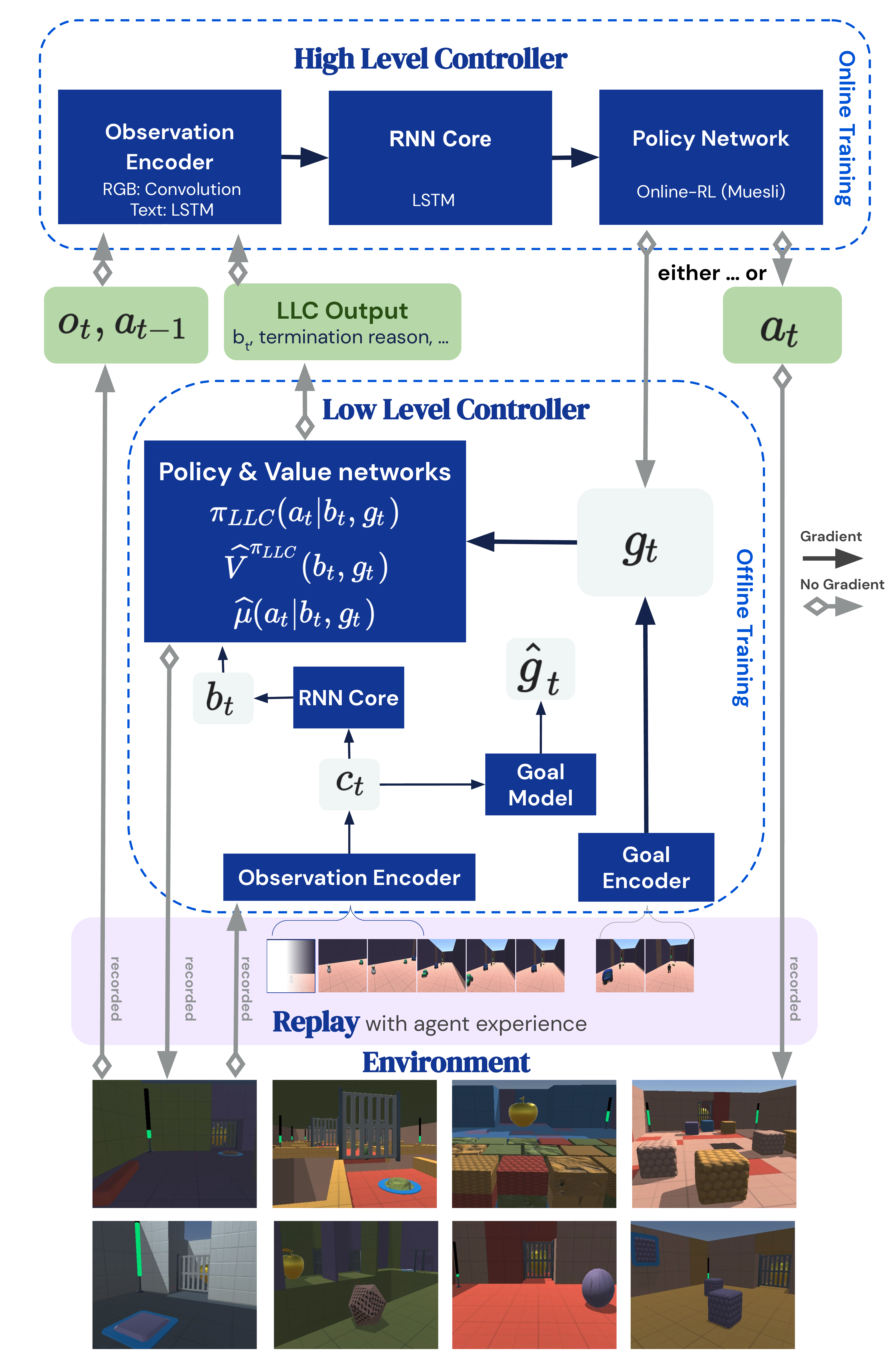}
    \caption{H2O2 component diagram. The dotted boxes indicate how components (in blue) are trained. The arrows indicate information (inputs, outputs and gradients) passed between components.}
    \label{fig:detailed_agent_architecture}
\end{figure}

\Cref{fig:detailed_agent_architecture} gives an overview of H2O2, and how its components interact with each other and the environment.
We outline H2O2's main components in the rest of this section, and we give details in \cref{sec:implementation_details}.

\subsection{Low-Level Controller Design}
\label{sec:llc}

\paragraph{Training data.}
The experience generated by H2O2 is inserted into a replay buffer \citep{horgan2018distributed,cassirer2021reverb} as the agent interacts with the environment.
The LLC learner processes minibatches of trajectories sampled (uniformly at random) from the replay \citep{horgan2018distributed}, where each trajectory has the form $(o_1, a_1, \ldots, o_n, a_n)$.

We sample \emph{start} and \emph{end} timesteps $t_s, t_e$ from the set $\{(i, j): 1 \leq i < j \leq n\}$, encode $o_{t_e}$ with the Goal Encoder (see \cref{fig:detailed_agent_architecture}) to obtain a \emph{latent goal} $g \in [-1, 1]^d$. 
The sampled goal is fixed for the sampled subtrajectory ($g_{t_s} = g_{t_s + 1} = \cdots = g_{t_e} = g$) and
each subtrajectory is treated as a separate episodic task terminating on timestep $t_e$ (when the goal $g$ is attained). 
The reward is $r_{t} \doteq \mathbb{I}\{t = t_e\}$ ($\mathbb{I}$ denotes the indicator function), that is, one for attaining the goal, and zero otherwise.
During training we sample multiple pairs $t_s, t_e$ per trajectory in the minibatch, so we train the LLC on multiple tasks (goals) at once.

The LLC's policy and value function are conditioned on the latent goal $g_t$ and on the agent's \emph{representation} $b_t$ \citep[the ``agent state''][]{sutton2018reinforcement}.
To compute $b_t$, we process the observations and actions using a recurrent IMPALA-like network \citep{espeholt2018impala} (the Observtion Encoder and the RNN Core in \cref{fig:detailed_agent_architecture}). 
For each $t$, the representation depends only on the previous observations and actions, and on the recurrent state of the neural network before $o_1$.

\paragraph{Goal Sampling during Training.} 
During goal sampling, we reject subtrajectories that are too short or too long (determined by hyperparameters), as well as goals that are too ``similar'' to other goals (details in \cref{sec:goal-sampling}). 
We also increase the relative frequency of pairs $(t_s, t_e)$ where the reward from the environment is non-zero for some timestep $t_s \leq t \leq t_e$.
Increasing the frequency of reward-related goals allowed us to direct the behavior of the LLC to be meaningful for the RL problem, without introducing too much of a dependency on task rewards, rather than learning options that directly optimize for the environment reward \citep{hafner2022deep}.
Controlling the goal sampling distribution also provides a direct way to study the option discovery problem using H2O2.

\paragraph{Offline RL.}
To learn the LLC policy $\llc$, we introduce a \emph{regularized} offline V-Trace \citep{espeholt2018impala,mathieu2021starcraft}.
$\llc$ is a distribution over primitive actions that we optimize by following a V-Trace policy gradient.
We also regularize $\llc$ to stay close to an estimate of the behavior policy $\widehat{\mu}$, trained with behavior cloning \citep{pomerleau1989alvinn}.
Similar to other offline RL works \citep{fujimoto2019offpolicy,gulcehre2021regularized}, we found this regularizer to be essential for training an effective LLC, as removing it or annealing it out made the offline RL ineffective.

The gradient step of regularized offline V-Trace is:
\begin{equation}
\begin{split}
    &\frac{\llc(a_t|b_t, g)}{\widehat{\mu}(a_t|b_t, g)} \cdot \Adv_t \cdot \grad \log \llc(a_t|b_t, g) \\
    &-\alpha \grad KL(\llc(b_t, g) | \widehat{\mu}(b_t, g)),
    \label{eq:llc-vtrace}
\end{split}
\end{equation}
where $\Adv_t$ is the advantage estimate at time $t$ (computed using V-Trace returns and the value estimate $\widehat{V}^{\llc}$, \citealp{espeholt2018impala}),
$\alpha$ is a fixed hyperparameter for the KL regularizer, 
and the gradient is taken only with respect to the parameters of $\llc$ (without differentiating through $\widehat{\mu}$).
Differently from the original V-Trace, \cref{eq:llc-vtrace} uses an estimate of the behavior policy instead of the behavior policy itself.

Following \citet{espeholt2018impala}, we add a weighted neg-entropy regularizer ($-H(\llc)$) to the objective for $\llc$.
We train the value function estimate $\widehat{V}^{\llc}$ through regression (as done by \citealp{espeholt2018impala}, and akin to Fitted Q Iteration, \citealp{ernst2005tree}).
The representation $b_t$ and the latent goal $g_t$ are shared between $\llc$, $\widehat{V}^{\llc}$ and $\widehat{\mu}$, and all three learning tasks (regularized policy gradient, value learning and behavior cloning) flow gradients into the representation and the latent goal.

\paragraph{Variational Goal Encoder.}
\label{sec:agent_kl_regularizer}
Our Goal Encoder is inspired by variational autoencoders \citep{kingma2013auto,rezende2014stochastic}
and it outputs a distribtuion over the goal space from which we can sample goals $g$.
Concretely, Goal Encoder outputs the parameters of a multivariate normal distribution with diagonal covariance that is used to sample the latent goal $g$.
Differently from VAEs, we do not attempt to autoencode the input of the Goal Encoder from the sampled latent goals, but instead use the sampled latent goals for the Offline RL and auxiliary tasks.

We use a KL regularizer term with weight $\beta$ to encourage the multivariate normal to stay close to a standard normal distribution.
A weight $\beta=0$ will allow the goal space to be as expressive as afforded by the Goal Encoder, whereas a large enough $\beta$ will effectively cause the goals $g$ to be sampled from a standard normal distribution (ignoring the observation input to the encoder).
The KL regularization is primarily for the benefit of the HLC, as our empirical results will show.
We believe that this regularization makes the goal space smoother, and make it easier for the HLC to explore and choose goals.

\paragraph{Auxiliary Tasks.}
In addition to the learning objectives outlined above,
we employed three auxiliary prediction tasks to improve the quality of our LLC.

The first one is training the Goal Model in \cref{fig:detailed_agent_architecture}.
via maximum likelihood estimation of latent goals $g_t$ conditioned on $c_t$ (the output of the agent's Observation Encoder).
This is an auxiliary only flow gradients into the observation encoder (not $g$).
We observed that, without this auxiliary task, the LLC would frequently ignore the goal conditioning.

The second auxiliary task is to estimate the state value of the behavior policy, with respect to the environment rewards. 
The value estimate is a function of $b_t$ and $g$ and flows gradient into both.
We found this auxiliary task to be beneficial, and we believe it helps by shaping $b_t$ and $g$ to encode information about rewarding states in the environment.

The third auxiliary task is to predict, from each timestep $t$, how far in future the goal is.
We frame the prediction task as a multiclass logistic classification task.
During training, if the goal is on step $t_e$, then the classification label for each step $t$ is $t_e - t$, out of $n$ possible classes, where $n$ is an upper-bound on how far into the future goals can be.

\subsection{SMDP and High-Level Controller Design}
\label{sec:hlc}

The HLC can instruct the LLC to execute either primitive actions or goals 
(and the decision of which one to choose at each step is part of the HLC's policy),
and the LLC executes them in the call-and-return model \citep{dayan1992feudal}.

\paragraph{Option Termination and Initiation.}
We improved our agent's sample efficiency by composing simple fixed rules and learned termination/initiation functions.
We used a hard limit on option duration \citep[timeout;][]{sutton1999between} as the fixed rule, and an ``attainment classifier'' as the learned termination function.
We built the attainment classifier from the LLC's auxiliary task that predicts (via classification) how far in the future the goal is.
The option terminates when class $0$ (``the goal is attained now'') is the most likely class predicted by the time-to-goal classifier.

The fixed initiation criterion is to allow any goal in any situation \citep{bacon2017option}.
However, this is problematic with learned goal-conditioned behavior because it is possible to task the LLC with attaining goals that cannot be attained---either because the goals are implausible, or because the LLC is incapable.
When the HLC requests an unattainable goal, the LLC will likely run until timeout, which has a significant cost in environment interactions.
We observed this to be very problematic in early training, as the HLC can frequently select unattainable goals, but is oblivious of the sample cost of doing so.

We addressed this issue by terminating options after one step if the goal was unattainable.
A goal was deemed unattainable if the value estimate $\widehat{V}^{\llc}(b_t, g)$ was below a certain threshold.
For a high enough threshold, this is a conservative criterion because the value estimates $\widehat{V}^{\llc}(b_t, g)$ will often only be high for goals that the LLC can achieve. 

\paragraph{HLC Observations.}
The LLC is responsible for what the HLC observes---it may forward environment observations, and it may also process and combine what has been observed during the execution of an option.
In this work, the LLC simply forwards the environment observation to the HLC on the steps where the HLC gets to observe the SMDP and take action.

\paragraph{High-Level Controller.}
We used Muesli \citep{hessel2021muesli} as the general-purpose RL algorithm for the HLC.
Muesli is a strong RL agent, and it is among the strongest RL agents in the Atari benchmark \citep{bellemare2013arcade}.
Moreover, it admits policies over continuous and discrete actions, and this allows us to parameterize the policies we need for interacting with the LLC. For additional implementation details see Appendix~\ref{sec:implementation_details}.

\section{Experiments}
\label{sec:experiments}

\subsection{H2O2 is competitive with a strong flat baseline}
\label{sec:experiments:performance}

We evaluated H2O2 in the DeepMind Hard Eight suite \citep{gulcehre2019making}.
These tasks are 3D, partially observable, procedurally generated, and they require exploration and complex behavior to solve (see \cite{gulcehre2019making} for detailed task descriptions).
The agents are trained in a multi-task setting in all eight tasks.
The flat baseline is a Muesli agent, and it has only the minimal, necessary differences from the HLC's Muesli agent---for example, the action spaces differ between H2O2 and the flat baseline, so the policies need to be changed accordingly.
Unless otherwise stated, all quantities reported in this work are binned over number of frames, and averaged across all tasks and over five independent runs.
Bands show standard error over independent runs.

\begin{figure}[htb!]
    \centering
    \includegraphics[width=0.90\columnwidth]{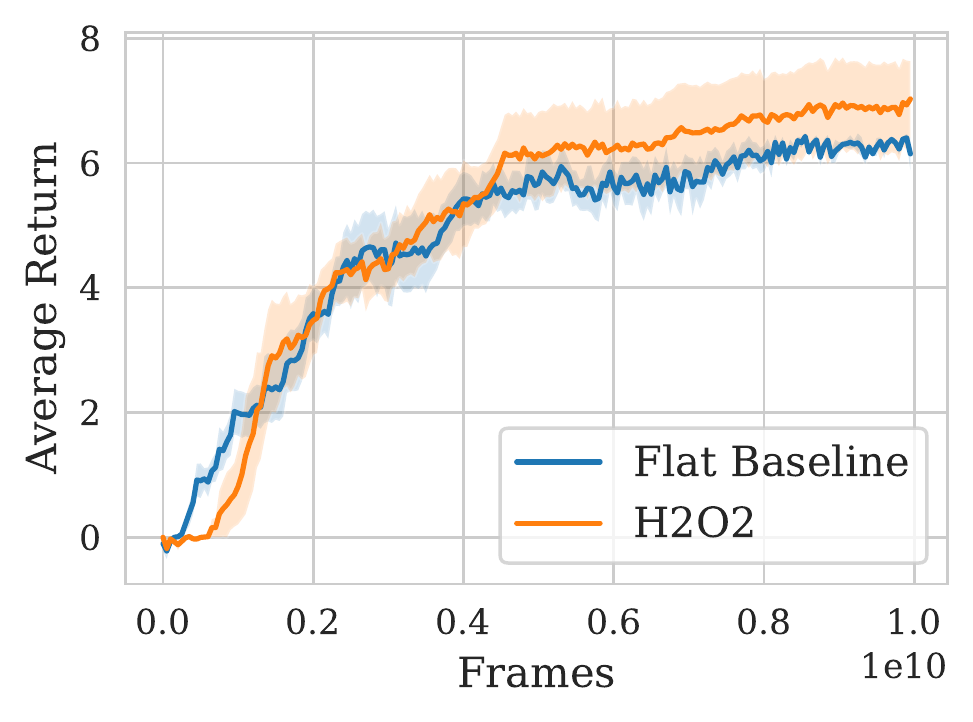}
    \caption{Average episode return for H2O2 and the Muesli baseline.
    }
    \label{fig:mean_return_main}
\end{figure}

\Cref{fig:mean_return_main} shows the average return per episode of H2O2 and the flat baseline as a function of the number of frames generated by interacting with the environment (i.e., \emph{throughout training}).
The plot shows that the two agents are competitive, with H2O2 attaining slightly higher performance more frequently.
We report per-task performance in \cref{fig:mean_return_per_episode_main_all_variants} in \cref{sec:per-level_performance},
where we can see different variations of sample efficiency and final performance between the two agents across tasks.

H2O2's improved performance is a demonstration of the effectiveness of our hierarchical agent, but the variations between H2O2 and the flat baseline performances in each task suggest that H2O2 is indeed learning differently from the flat baseline.
How is the hierarchical design influencing H2O2's learning and final performance?
What HRL capabilities is H2O2 demonstrating?

\subsection{Is H2O2 using temporally extended behaviors?}
\label{sec:experiments:temporal_abstraction}

Yes, but we did not observe that ``the more temporal extension, the better''.
We found that parameters that control temporal extension have to be carefully selected in order to obtain better performance and even, ``paradoxically'', temporally extended behavior.
The apparent paradox stems from considering the benefits of increasing temporal extension, without accounting for how it impacts the learning problem.
That is, an effective hierarchical agent with more temporally extended behavior is expected to perform at least as well as one with less temporally extended behavior,
but giving an untrained agent access to more temporally extended behavior may make the learning problem harder.
The problem may be so hard that even after significant training the learning agent may have subpar performance and it may
fails to display any meaningful hierarchical behavior.

To substantiate our claim, we measured the average number of environment (LLC) steps per SMDP (HLC) step, as well as task performance, of different variants of H2O2.
Higher values of ``average LLC steps'' (per HLC step) mean H2O2 spent more timesteps in temporally extended behavior.
An agent that exclusively executed primitive actions would have a ratio of $1$.
This ratio allows us to infer the fraction of steps spent executing options excluding the first step, which for the purpose of our discussion is how much temporally extended behavior an agent displays.

The typical range for the the average LLC steps is between $1.0$ and $2.5$.
An average LLC steps ratio of $1.5$ means the agent is in temporally extended behavior for about $\frac{1}{3}$ of its interaction with the base environment.
A ratio of $1.25$ corresponds to temporal extension in at least $20\%$ of the interaction, and a ratio of $2.5$ is corresponds to at least $60\%$.

\begin{figure}[htb!]
    \begin{subfigure}[t]{0.49\columnwidth}
       \centering
    \includegraphics[width=\columnwidth]{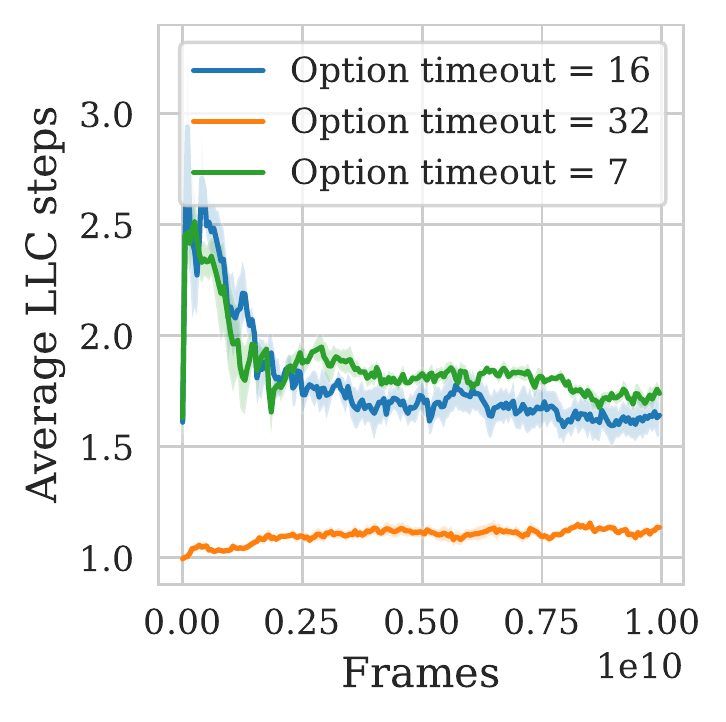}
    \caption{\label{fig:mean_llc_steps_ablation}}

    \end{subfigure}
    \centering
    \captionsetup[subfigure]{justification=centering}
    \begin{subfigure}[t]{0.49\columnwidth}
   \centering
    \includegraphics[width=\columnwidth]{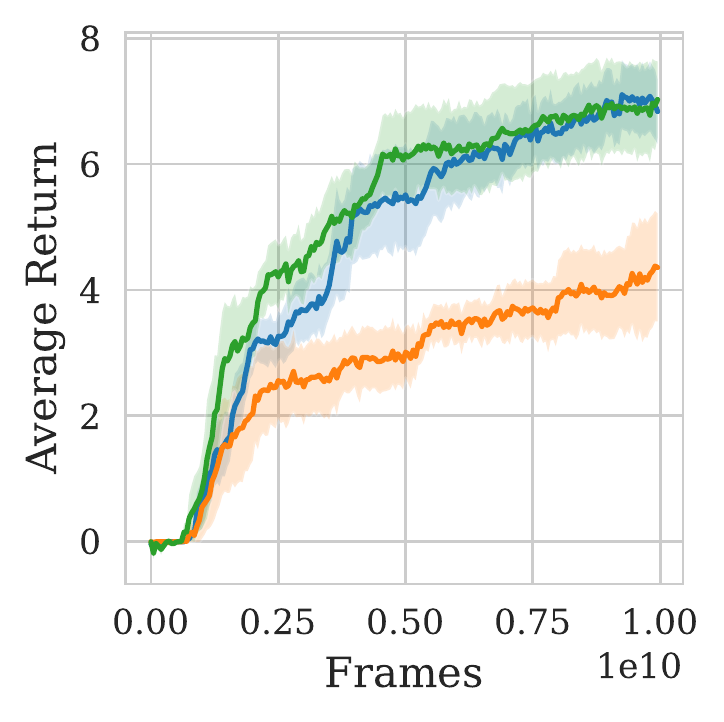}
    \caption{\label{fig:max_option_duration_ablation}}
    \end{subfigure}
    \captionsetup[subfigure]{justification=centering}
    \caption{Evaluation of H2O2 with different option timeouts. 
    (\subref{fig:mean_llc_steps_ablation})
    Average number of LLC steps per HLC steps throughout training
    All curves are at approximately $1.0$ in the initial training steps; the apparent start of some curves at higher values is an artifact of binning.
    (\subref{fig:max_option_duration_ablation}) Average return of H2O2 for different option timeouts. }
    \label{fig:max_option_duration}
\end{figure}

\paragraph{H2O2 with different option timeouts.}
We considered variants of H2O2 with different timeouts: $7$, $16$ and $32$ steps (the timeout used for H2O2 in \cref{fig:mean_return_main} was $7$).
Options terminate when either the goal is attained (according to the LLC's classifier) or at timeout.
Because the options have a termination function, we would expect that increasing timeout should increase the effectiveness and frequency of the agent's temporally extended behavior.
\Cref{fig:mean_llc_steps_ablation} shows, however, that this is not the case.
Surprisingly, attempting to increase the amount of temporal abstraction by increasing timeouts eventually \emph{harms} H2O2's ability to employ temporally extended behavior.
Moreover, H2O2's performance is surprisingly sensitive to the amount of temporal abstraction: 
even with a timeout of $16$ (which has roughly the same amount of temporally extended behavior (\cref{fig:mean_llc_steps_ablation}) as the timeout of $7$),
H2O2's performance is worse than with a timeout of $7$ (see \cref{fig:max_option_duration_ablation}).
The performance with the timeout of $32$ is worst, so this setting leads to poor behavior both in terms of temporal abstraction and task performance.

Our data suggests that H2O2 with a timeout of $32$ breaks down because the learning problem is too hard.
We measured why and how often options terminate throughout training (see \cref{fig:max_llc_steps_32_termination_reason}), and we saw that in this setting our agent spends about half of the first $2 \cdot 10^9$ frames issuing invalid goals (and thus generating transitions with no-ops).
So we suspect that the HLC failed to generate ``good'' data for the LLC to learn effective goal-following behavior, which in turn led to an unnecessarily challenging SMDP for the HLC to solve, that is, one filled with useless options that waste several frames of environment interaction.

\begin{figure}[htb!]
    \begin{subfigure}[t]{0.49\columnwidth}
        \centering
        \includegraphics[width=\columnwidth]{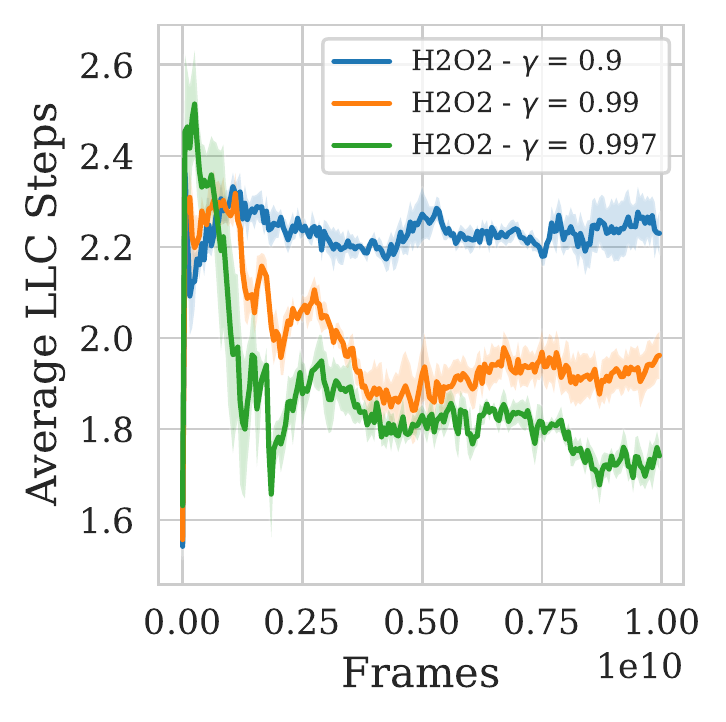}
        \caption{\label{subfig:llc_steps_discount_ablation}Average number of LLC steps per option.}
    \end{subfigure}
    \centering
    \captionsetup[subfigure]{justification=centering}
    \centering
    \begin{subfigure}[t]{0.49\columnwidth}
        \centering
        \includegraphics[width=\columnwidth]{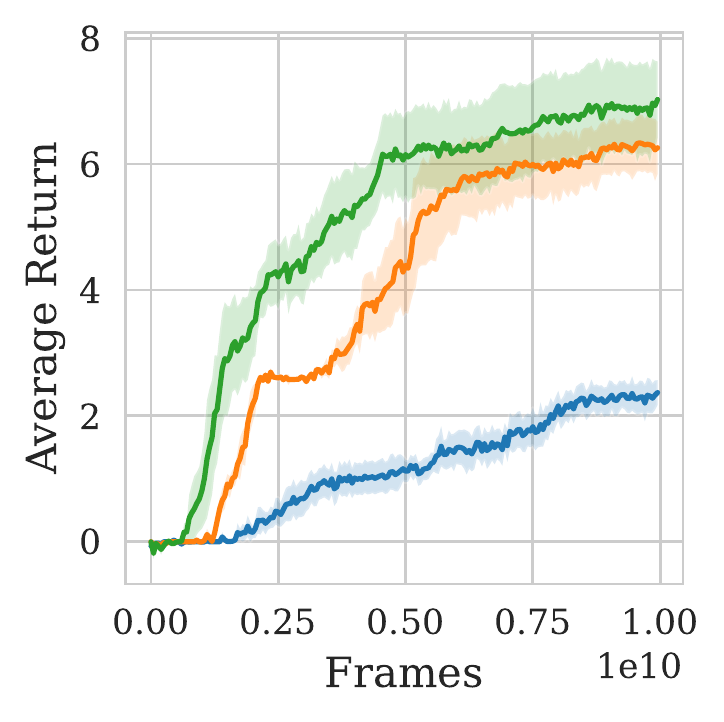}
        \caption{\label{subfig:return_discount_ablation}Average return.}
    \end{subfigure}
    \captionsetup[subfigure]{justification=centering}
    \caption{Option use (left) and performance (right) for H2O2 with different discount factors $\gamma$.}
    \label{fig:discount_ablation}
\end{figure}

\paragraph{H2O2 with different discounts.}
We also considered three variants of H2O2 with different discounts $\gamma$, in $\{ 0.9, 0.99, 0.997 \}$ (the value used for H2O2 in \cref{fig:mean_return_main} was $0.997$).
Since the discounting only incides on HLC timesteps, the rewards $n$ timesteps in the future will only be discounted by $\gamma^n$, even if the amount of primitive actions required to get to that state is significantly larger.
For example, if options take an average of $1.5$ steps, $\gamma = 0.9$ and $n = 10$, the reward for HLC would be discounted with $0.59$, whereas a flat agent executing the same actions would see the reward discounted with $0.21$.
This ``horizon shortening'' is expected to encourage the agent to use options, so we expect to see the variants with smaller $\gamma$ using more options.
\Cref{subfig:llc_steps_discount_ablation} shows that this is indeed the case.

\Cref{subfig:return_discount_ablation} shows that as we decrease $\gamma$ the agent spends more time in temporally extended behavior.
HRL folklore suggests that the agent with more temporally extended behavior will perform better, because it will be able to assign credit at longer timescales.
The results in \cref{subfig:return_discount_ablation} are evidence against this claim: The plot shows that H2O2 with the largest $\gamma$ performs best, and that decreasing $\gamma$ worsens perfomance (even though it increases option use, as shown in \cref{subfig:llc_steps_discount_ablation}).
The issue is that changing $\gamma$ also affects the objective of the HLC, and that changing the objective can change both the final solution and how the agent explores.
Our hypothesis is that H2O2 with lower $\gamma$ explores worse, possibly in three ways: 1) The HLC fails to generate behavior with higher rewards because it is optimizing for short-term cumulative reward; 2) The options learned from the poor-performing HLC are also poor (with respect to the task), and the agent is incentivized to choose poor-performing options over exploring further with primitive actions; 3) The longer options also reduce the amount of training data the HLC generates for itself; that is, the HLC is incentivized to generate less data for itself by using options. 
This third point makes the agent sample inefficient!

\Cref{sec:action_repeat} shows that H2O2 outperforms a simpler idea of adding temporal abstraction to a flat baseline by increasing the number of repeated actions.

\subsection{Does H2O2 benefit from more options?}
\label{sec:experiments:option_space}

Sometimes, and we argue that it depends on whether the options \emph{simplify the problem}.

HRL folklore suggests that making more skills available to the HLC empowers the agent and leads to better solutions.
We claim that this is not necessarily the case, and that in H2O2, for learned options to be beneficial, they must \emph{simplify} the problem for the HLC.
That is, it does not make sense to solve an SMDP that is harder to solve than the original MDP---in that case we are better off using the flat agent.
We considered two ways to offer more options to H2O2: Increasing the dimension of the latent goals, and reducing the amount of regularization on the goal space.

\begin{figure}[htb!]
    \centering
    \captionsetup[subfigure]{justification=centering}
    \begin{subfigure}[t]{0.49\columnwidth}
        \centering
        \includegraphics[width=\columnwidth]{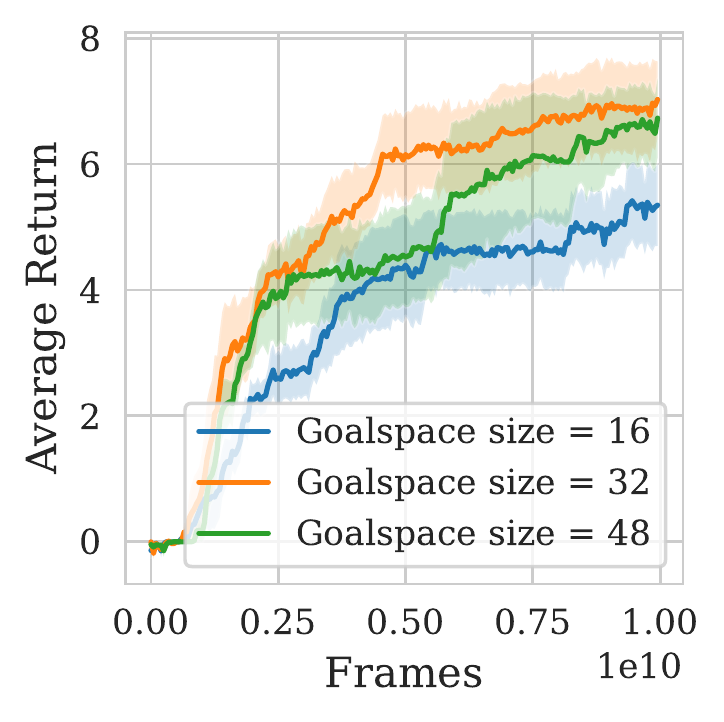}
        \caption{Average return for H2O2 with different latent goal dimensions.\label{fig:goalspace_size_ablation}}
    \end{subfigure}
    \centering
    \begin{subfigure}[t]{0.49\columnwidth}
        \centering
        \includegraphics[width=\columnwidth]{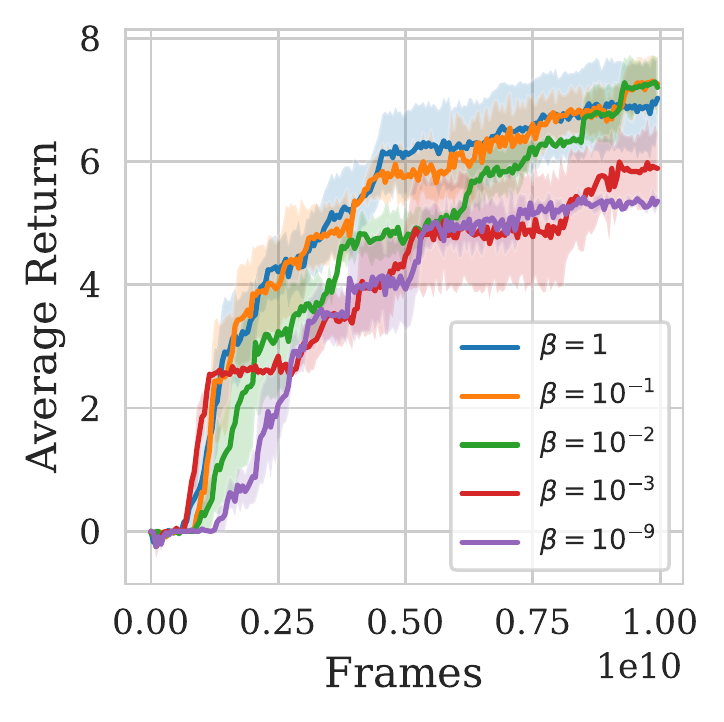}
        \caption{Average return for H2O2 with different amounts of goal compression. The higher $\beta$, the more compression there is.\label{fig:kl_ablation}}
    \end{subfigure}
    \caption{Effect of constraining the space of options on overall performance.}
\end{figure}

\Cref{fig:goalspace_size_ablation} shows the performance of H2O2 where we varied the dimension of the latent goals. 
We considered dimensions in $\{16, 32, 48\}$, and H2O2 from \cref{fig:mean_return_main} uses $32$.
We see in the figure that this dimension behaves like a usual hyperparameter: There's a sweetspot for highest performance at $32$, but going lower or higher leads to worse performance.
It is surprising, though, that a dimension of $32$ works best, as we were initially expecting more expressive goals to be more effective.

We also evaluated the effect the Goal Encoder regularizer on the performance of H2O2.
We considered $\beta \in \{10^{-9}, 10^{-3}, 10^{-2}, 10^{-1}, 1\}$.
When $\beta = 10^{-9}$ there is no compression of the goal space, but when $\beta = 1$ the regularization is so strong that the posterior is also a standard normal.
We used $\beta = 1$ for H2O2 in \cref{fig:mean_return_main}.

\Cref{fig:kl_ablation} shows the performance of H2O2 for the different values of $\beta$,
and we see that H2O2 with the least diverse set of options ($\beta = 1$) performs best,
along with the larger values of $\beta$.
The data is consistent with the hypothesis that too much flexibility in the goal space makes the learning problem harder, so adding more options eventually damages the performance of the agent.

\subsection{How does H2O2 perform in similar domains from related work?}

The work of \citet{hafner2022deep} is the closest to ours: They introduced Director, a hierarchical deep RL agent for complex partially observable, first-person 3D environments.
The was evaluated in \texttt{Goals Small} and \texttt{Objects Small} in the DeepMind Lab environment \citep{beattie2016deepmind}.
These are first-person maze tasks that require navigation, localization and recalling reward locations within an episode.
Director is an SMDP agent with call-and-return options, but without access to primitive actions, and all options terminate after a fixed number of steps.
The options are ``state''-directed, in the sense that the LLC (``Worker'') is conditioned on latent vectors from the latent state space (the analogue of our $b_t$ in \cref{fig:detailed_agent_architecture}), and is trained in hindsight for achieving the corresponding latent states.
\citet{hafner2022deep} use a VQ-VAE \citep{van2017neural,razavi2019generating} to discretize the latent state space, which gives the HLC (''Manager'') a discrete action space.
Moreover, they use a World Model \citep{hafner2019learning} to help shape Director's state representation.
During training, the LLC in Director is rewarded proportionally to the similarity of its latent state and the conditioning input (the goal), at each timestep.
The version of Director that is competitive with their baseline \citep[Dreamer;][]{hafner2019dream} adds extrinsic rewards to the reward provided to the LLC.

\begin{figure}[htb!]
    \centering
    \includegraphics[width=\columnwidth]{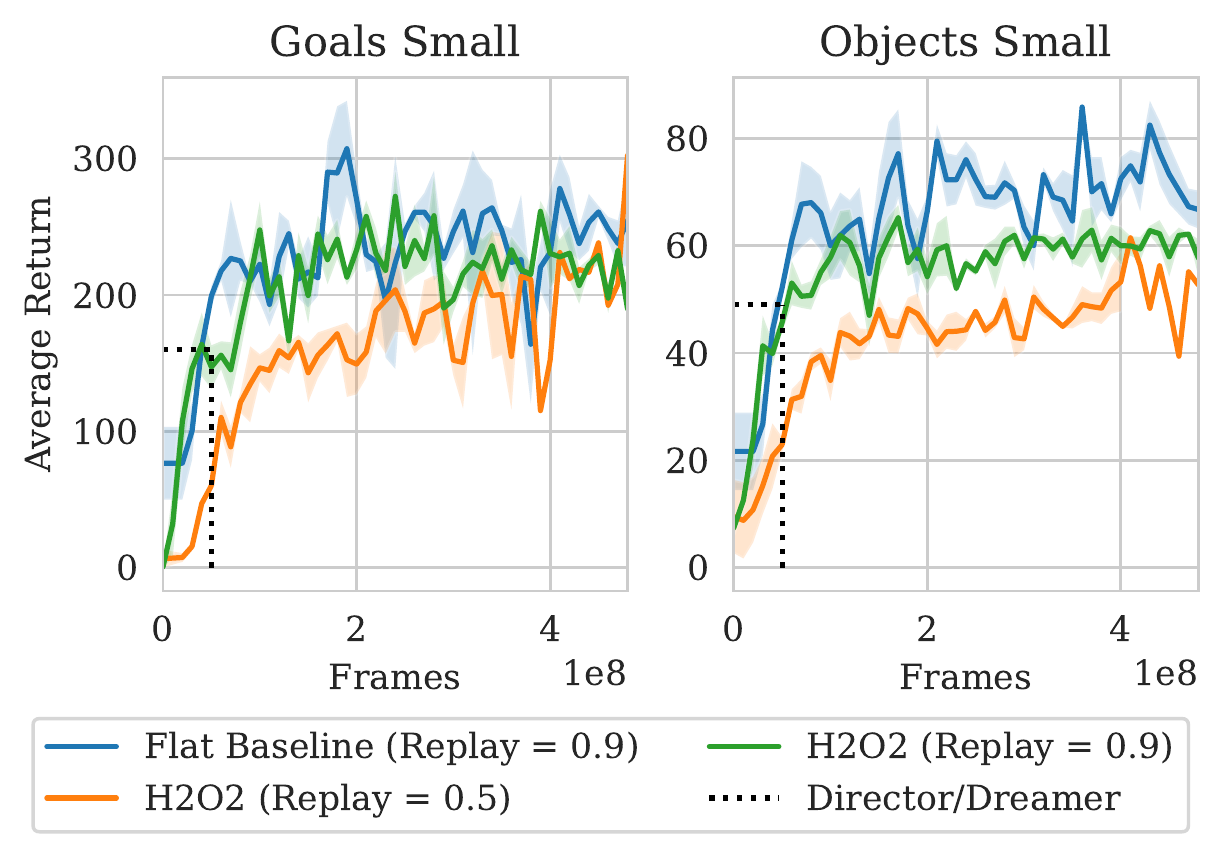}
    \caption{The average return across two levels of DMLab \cite{beattie2016deepmind}. We also indicate the final performance of the Dreamer and Director baselines \cite{hafner2022deep} with the dotted line, after 50M frames.}
    \label{fig:dmlab}
\end{figure}

We evaluated H2O2 and our flat Muesli baseline in DeepMind Lab's \texttt{Goals Small} and \texttt{Objects Small}.
\Cref{fig:dmlab} shows the average return of different agents on the two tasks.
The ``Flat Baseline'' is a Muesli agent like the one used in the Hard Eight tasks, but uses a replay-to-online ratio of $0.9$\footnote{This ratio means that in each minibatch $90\%$ of the data is sampled from a replay, and the other $10\%$ from online experience. This increases the data efficiency of the agents and make them competitive in early training (the $180$-thousand-frame regime).}
We present variants of H2O2 with the replay-to-online ratio used for Hard Eight tasks ($0.5$) as well as $0.9$.
The figure also shows the final performance of Director and Dreamer (as the dotted line, both methods have the same final performance).
This variant of Director adds extrinsic rewards to the LLC objective.
\Cref{fig:dmlab} shows that with an appropriate replay-to-online ratio both H2O2 and Muesli baseline can match the data efficiency of Director and Dreamer, though it's unclear what the latter's final performance would be if trained longer.

\section{Conclusion}

Our work introduces H2O2, the first demonstration of a complete hierarchical agent that can use non-trivial options to attain strong performance in visually complex partially observable tasks.
H2O2 is competitive with a state-of-the-art flat baseline, it discovers and learns to use its options, and it does so from its own generated experience. 

\paragraph{Relevance.}
HRL has received much interest due to its potential to deliver powerful agents with appealing capabilities---for example, transfer, skill reuse, planning and exploration with abstraction.
Recent successes with HRL in different domains \citep{merel2018neural,wulfmeier2021data,hafner2022deep} provide evidence that practical, effective HRL agents are possible, even if existing agents do not yet fully realize the potential of HRL.
Therefore, it is important to expand the coverage of effective hierarchical agents across domains,
and to identify and tackle practical challenges that can bring us closer to a full-fledged hierarchical agent.

\paragraph{Significance.} Our work is an important contribution to the HRL research for two reasons: H2O2 is a proof of concept, complete and effective HRL agent, and our work highlights critical challenges for HRL in complex domains that are vastly overlooked in the literature.
It was only by going through the process of designing and training a HRL agent for complex domains that we exposed some of these issues.

\paragraph{Successes.} We built on existing work to tackle some of the practical challenges of HRL in complex domains, such as learning goal-conditioned behaviors offline from any experience generated by an agent (not jut expert behavior).
To achieve this, we introduced a regularized offline V-Trace algorithm and demonstrated how to integrate the policy that executes these goal-conditioned behaviors (the LLC) with a general-purpose RL algorithm that learns to select these behaviors as options in order to solve tasks (the HLC).

\paragraph{Lessons learned.}
We believe that our empirical findings apply to domains where a very large number of options is conceivable, but for any one task a much smaller set of behaviors is relevant and useful.
Visually complex domains tend to naturally have this property, and this is arguably the kind of rich domain we want intelligent agents to be effective in.
However, we think that many of the challenges we observed would go away if we were to limit the learning to a small set of options \citep{merel2018neural,wulfmeier2021data}, or choose them sensibly  beforehand.

Within the scope of these ``rich domains'', however, the lessons we can draw from our experimental results can apply to various HRL agents beyond H2O2.
The lessons apply most closely to SMDP agents.
The SMDP framework has been backed with theoretical justification\citep{precup2000temporal,barto2003recent}, and our work complements existing knowledge with empirical findings.

We noticed a strong contrast between how HRL is typically motivated in the literature \citep[e.g.,][]{barto2003recent,pateria2021hierarchical,hutsebaut2022hierarchical}, and the practical challenges we encountered.
It is often claimed that hierarchical agents can demonstrate very appealing capabilities and algorithmic strengths, such as sample efficiency, structured exploration, temporal abstraction, improved credit assignment, state abstraction, jumpy planning, transfer and generalization.
These ``HRL promises'' can easily be misconstrued as properties of hierarchical agents, which may lead to misconceptions about how hierarchical agents will learn and perform.

Our empirical findings exposed some of these HRL misconceptions.
For example, the SMDP approach promises to simplify the problem for the general-purpose RL algorithm.
So one might expect that adding capabilities that are perceived as strengths of HRL (for example, more expressive options) to the SMDP will cause the general-purpose RL algorithm to solve the SMDP with less effort than if it had simpler options, or only primitive actions.
However, in some experiments we showed the opposite.

In practice, the design of the LLC effectively changes the SMDP, and the hierarchical agent can only be competitive with a flat agent if SMDP is easier to solve than the original MDP (besides admitting a better solution).
Therefore both solution quality and learning dynamics are essential factors to consider when designing the hierarchical agent.

\paragraph{Open challenges.} 
We also identified questions that remain open: 
How can we structure the goal space to accelerate the HLC learning? 
Is it possible to learn effective HLCs with a general-purpose RL algorithms?
How can the HLC agent learn with a very large number of complex options, but remain competitive with a flat baseline?
Are image goals good enough?
What other goal modalities can we use?
Which goals should we train the LLC to achieve?

Some of these questions can be investigated in simple domains, as long as the domains are designed to pose challenges that we observe in practice.
For example, a simple grid-world where there is an option to reach any cell from any other cell can be a fruitful domain to explore.
However, it may be challenging to outperform strong flat deep RL baselines in such simple domains if the options are not prescribed but learned end to end.

We presented simple approaches for some of the challenges above---e.g.~the goal sampling distribution for the LLC.
We expect that the performance of H2O2 will improve with goal sampling distributions that incorporate principled techniques for option discovery \citep{machado2016learning}. 
H2O2 can be a starting point for research that aims to investigate specific HRL sub-problems without losing sight of the performance of the whole agent in complex tasks.

\bibliographystyle{abbrvnat}
\setlength{\bibsep}{5pt} 
\setlength{\bibhang}{0pt}
\bibliography{references}

\appendix

\section{SMDP Agent Design Details}

\begin{figure}[ht!]
    \centering
    \includegraphics[width=0.85\columnwidth]{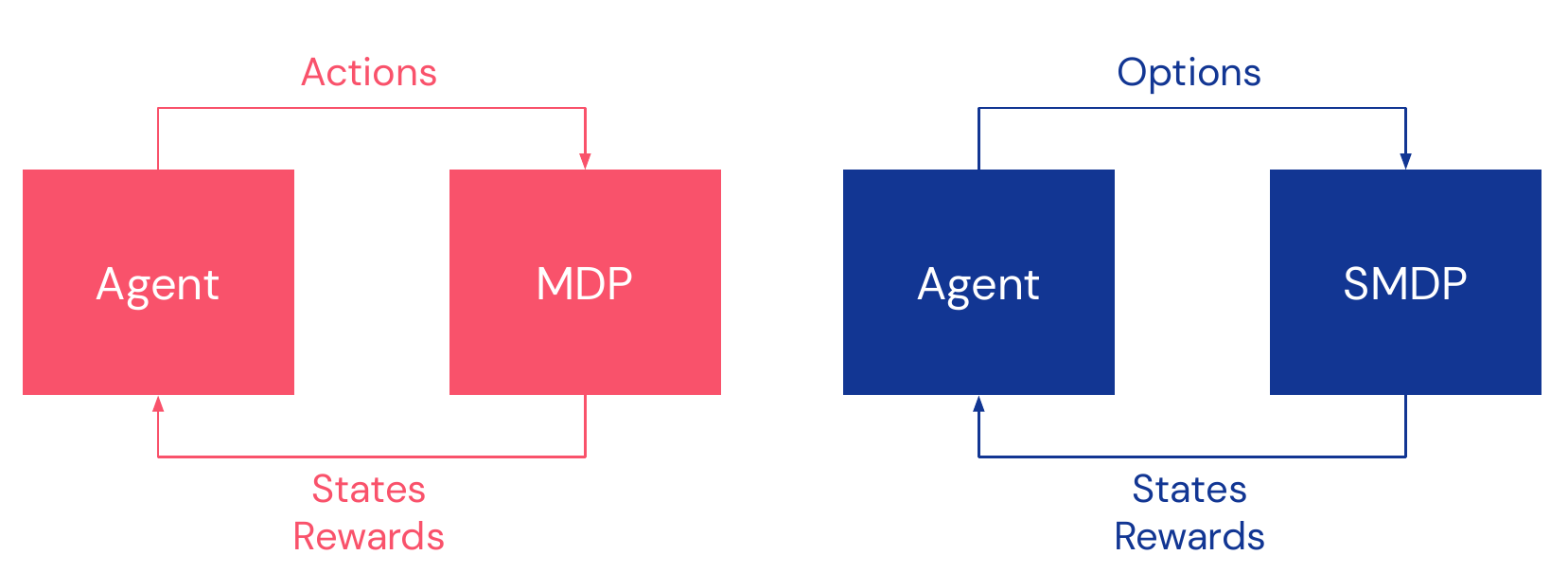}
    \caption{Agent-environment interaction diagrams for MDPs (left) and SMDPs (right).}
    \label{fig:mdp_smdp_diagram}
\end{figure}

\begin{figure}[ht!]
    \centering
    \includegraphics[width=0.9\columnwidth]{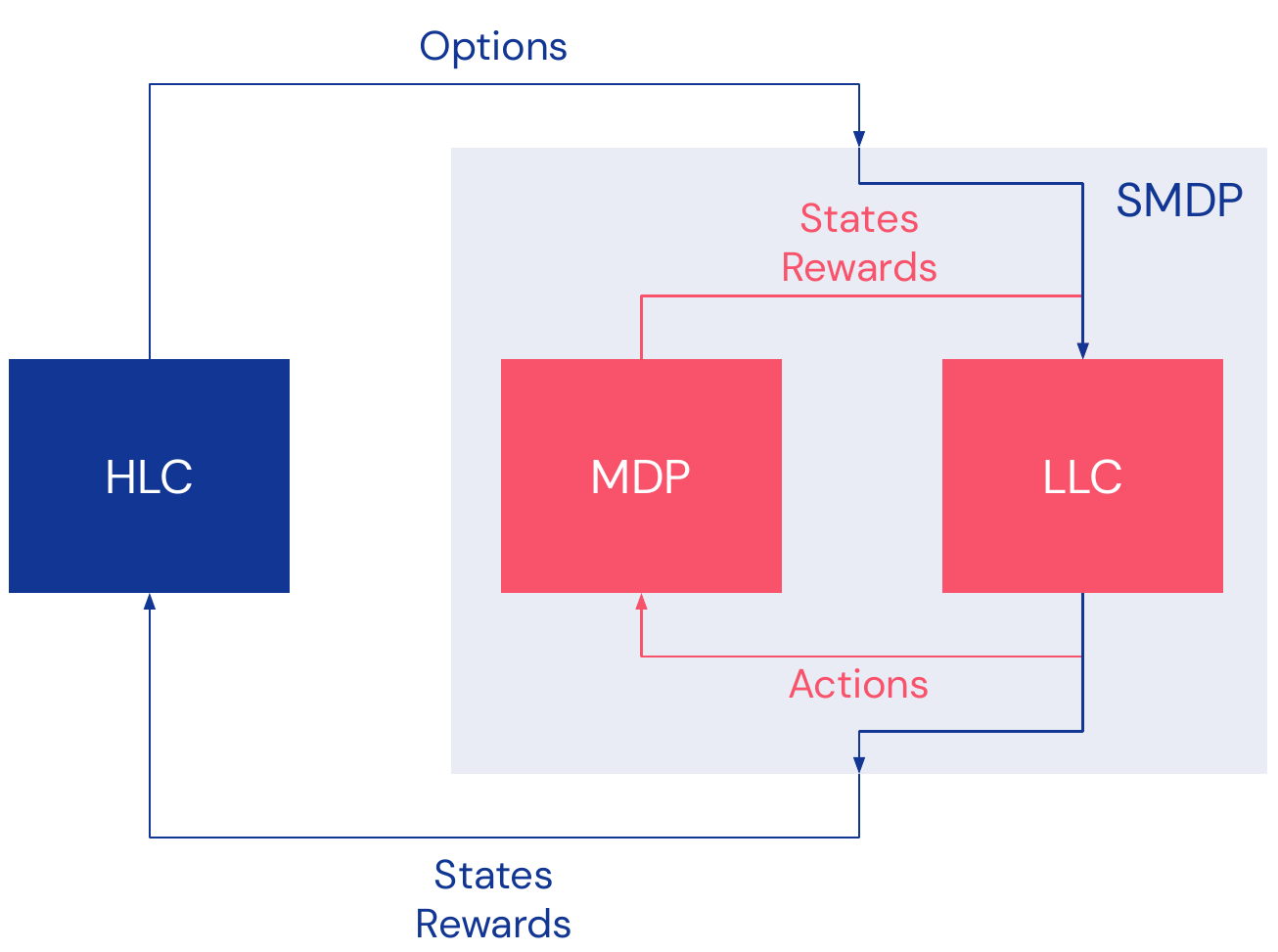}
    \caption{Agent-environment interaction diagrams for an SMDP, framed in terms of high-level and low-level controllers.}
    \label{fig:hlc_llc_diagram}
\end{figure}

An RL agent interacts with a semi-MDP \citep[SMDP;][]{sutton1999between} by selecting options (temporally extended behaviors with initiation and termination conditions), whereas an RL agent interacting with an MDP \citep{sutton2018reinforcement} selects (primitive) actions.
\Cref{fig:mdp_smdp_diagram} outlines these interaction loops, which are intentionally similar.
We can turn an MDP into an SMDP by adding options to it.
When we learn the options with RL, it is useful to structure the agent interaction in terms of a high-level controller (HLC) which interacts with the SMDP, and a low-level controller (LLC), which interacts with the MDP and provides the options for the SMDP.
\Cref{fig:hlc_llc_diagram} outlines the interaction loops for the LLC and the HLC.
The diagram corresponds to how we structured the HLC and LLC in H2O2.

The LLC interacts directly with the MDP (the agent-MDP interaction in the left diagram of \cref{fig:hlc_llc_diagram}, in red).
It is responsible for changing its behavior in response to the options received, and for providing the SMDP states and rewards for the HLC, based on the MDP states and rewards it observes.

The LLC executes options in the call-and-return model \citep{dayan1992feudal}: The HLC selects an option, then the LLC executes the option until it terminates, then returns control to the HLC.
From the perspective of the HLC, the interaction with the environment is an agent-SMDP interaction (the right diagram in \cref{fig:hlc_llc_diagram}, in dark blue).

In our work, the LLC allows the HLC to select primitive actions \citep{sutton1999between}, which are implemented as one-step options where the LLC executes the action requested by the HLC.
Our LLC forwards environment observations to the HLC along with a ``termination reason''.
The termination reason is a flag indicating why the LLC execution terminated, and it can be one of the following:
\begin{itemize}
    \item First timestep: The first timestep in an MDP episode requires the HLC to select an action in the SMDP for the LLC to perform.
    \item Last timestep: The last timestep in an MDP episode is also the last timestep in the corresponding SMDP.
    \item The option terminated. In this case the termination reason encodes why the option terminated. This could be because the goal was attained, the option timed out, or because it was a primitive action (which is framed as a one-step option).
    \item The option failed to initiate. This is because the goal corresponding to the option is deemed unattainable. The LLC executes a no-op in the environment and terminates. The forced no-op is an implementation decision to prevent the HLC from selecting unattainable goals indefinitely.
\end{itemize}

\section{Implementation details}
\label{sec:implementation_details}

\subsection{Goal Sampling Distribution}
\label{sec:goal-sampling}

We rejected ``similar'' goals as follows:
In a trajectory of potential goals $g_1, \ldots, g_n$, we first identify, for each candidate goal $g_t$, the closest that any other goal is:
\[
\mathrm{sim}_t \doteq \max_{t'} \frac{\langle g_t, g_{t'} \rangle}{\|g_t\|_2\|g_{t'}\|_2}.
\]
Then we exclude all pairs $(t_s, t_e)$ from sampling, such that $\mathrm{sim}_{t_e}$ is larger than the $40$-percentile of $(\mathrm{sim}_1, \ldots, \mathrm{sim}_n)$.

\subsection{Low-Level Controller}

\paragraph{Image Observations and Goals.}
Image observations and image goals were \texttt{72 x 96} RGB first-person views of the environment, encoded with a ResNet followed by an MLP \citep{hessel2019multi}.
For detailed information on the various model hyper-parameters, refer to \Cref{tab:llc-hyperparameters}.

\paragraph{Vector Goals and Goal Distributions for the affordance model.}
Throughout this work, we used a goal space defined as $\Goals = [-1, 1]^{32}$. We explained how the this goal space was modelled as a multivariate Normal with independent components in \cref{sec:llc}. 
For the affordance model on the other hand, we used a different parameterization of the goal distribution that is more flexible and expressive than a Normal distribution.   
Specifically, we parameterized the goal distributions with categorical random variables over the discretized $[-1, 1]$ range for each dimension, similar to how we parameterized the continuous actions.

\paragraph{Policy.}
The policy used in the LLC was the same as that used for the primitive actions in the HLC, which is that used in Muesli \citep{hessel2021muesli}.

\begin{table*}
  \begin{center}
    \caption{\label{tab:llc-hyperparameters} Training hyper-parameters for the LLC.}
    \begin{tabular}{ll}
    \toprule
    \textbf{Network} \\
    \midrule
    Image encoder (Observation) & ResNet with SAME padding  \\
    ResNet channels & $(16, 32, 32)$ \\
    ResNet residual blocks & $(2, 2, 2)$  \\
    ResNet kernel sizes & $(3, 3, 3)$ \\
    ResNet kernel strides & $(1, 1, 1)$ \\
    ResNet pool sizes & $(3, 3, 3)$ \\
    ResNet pool strides & $(2, 2, 2)$ \\
    MLP sizes &  $(512, 512)$ \\
    Proprioceptive encoder MLP sizes & $(256, 64)$ \\
    State LSTM size & $512$ \\
    Policy conditioning & Concatenation \\
    Policy LSTM size & $512$ \\ 
    Policy Head MLP shapes & $(512, 256, 128, 196)$  \\
    Number of components per continuous actions & $8$ \\
    Policy Distribution for continuous actions & Mixture of Clipped Logistics \\ 
    Policy Distribution for discrete actions & Categorical \\
    Value Head MLP shapes & $(512, 256, 128, 1)$  \\
    Number of actions & $10$ \\
    \midrule
    \textbf{Goal Encoder}\\
    \midrule
    Goal ResNet Encoder & Same as Observation ResNet \\
    Goal MLP sizes & $(512, 512, 256, 256, 32)$ \\
    Goal activation & LayerNorm + tanh activation \\
    \midrule
    \textbf{Auxiliary Losses}\\
    \midrule
    Behavioral Policy Head MLP & $(512, 256, 128, 196)$  \\
    Goal Model MLP & $(512, 256, 128)$ \\
    Goal Model Distribution & Binned bounded range per dim.\\
    Number of bins per dimension & $21$ \\
    External State Value MLP & $(256, 256, 1)$ \\
    Attainment Predictor LSTM size & $256$ \\
    Attainment Predictor MLP sizes & $(256, 256, 64)$ \\
    \midrule
    \textbf{Losses}\\
    \midrule
    V-Trace discount factor $\gamma$ & $0.8$\\
    $\widehat{V}^{\llc}$ objective weight & $0.5$ \\
    KL-Regularizer (to Behavior Policy) weight ($\alpha$ in \cref{eq:llc-vtrace}) & $10^{-2}$ \\
    Neg-entropy regularizer ($-H(\llc)$) weight & $10^{-3}$ \\
    $\rho$ clipping threshold & $1$ \\
    External state value discount factor & $0.99$ \\
    Optimizer & Adam \\
    Learning Rate & $10^{-4}$ \\
    Global Norm clipping & $10000$ \\
    Batch size & $16$ \\
    Unroll Length & $64$ \\
    Action repeat & $4$ \\
    \bottomrule
    \end{tabular}
  \end{center}
\end{table*}

\subsection{High-Level Controller}

For the HLC in this work, we used a Muesli agent with adaptations to the policy, so that it could select high-level actions (see \cref{sec:hlc}).
The primitive observations were processed as in the original Muesli agent, and the LLC observations were treated as scalar and vector observations and processed like other scalar and vector observations in Muesli---that is, input to an MLP and the result concatenated to the other processed observations. For specific hyper-parameters different from those used by \citet{hessel2021muesli}, refer to \Cref{tab:hlc-hyperparameters}.

\paragraph{Vector Goal Policy.}
To select vector goals in the HLC, we augmented the action space of the Muesli agent \citep{hessel2021muesli} with 32 extra continuous actions, the same dimension as our continuous goal space $G \in \Goals$. We also added a Bernoulli distribution for choosing between primitive actions and options as well as additional actions for the LLC hyperparamters set by the HLC, as described in \cref{sec:hlc}.

\begin{table*}
  \begin{center}
    \caption{\label{tab:hlc-hyperparameters} Training hyper-parametersfor the HLC and Flat Muesli baseline. The only difference is the number of actions, where for the baseline there are only the primitive actions. For parameters we do not provide values to, refer to \citet{hessel2021muesli}.}
    \begin{tabular}{ll}
    \toprule
    Replay proportion in batch & $0.5$ \\
    Discount factor & $0.997$ \\
    Optimizer & Adam \\
    Learning rate & $3 \cdot 10^{-4} $ \\
    Model unroll length & $2$ \\
    Policy weight & $3$ \\
    Auxiliary weight & $1$ \\
    Policy Distribution for discrete actions & Categorical \\
    Policy Distribution for continuous actions & Mixture of Clipped Logistics \\
    Number of components per continuous action & $8$ \\
    Number of actions & $43$, factored as \\
    &$\quad 1$ (execute option or primitive action, binary)  \\
    &$\quad 32$ (goal, continuous) \\
    &$\quad 10$ (primitive action, $8$ continuous, $2$ binary) \\
    Action repeat & $4$ \\
    \bottomrule
    \end{tabular}
  \end{center}
\end{table*}
\section{Additional Results}
\label{sec:additional_results}

\subsection{Exploring the HLC-LLC Interface and per-level performance}
\label{sec:per-level_performance}

It was very surprising to find out that the best H2O2 performance came from the agent with maximum goal space compression, so we wanted to further investigate whether completely removing the goal space and allowing the LLC to execute an unconditional reward-maximising policy would perform at a similar level. The hypothesis is that since the goal space is strongly regularized towards a standard Normal, the \emph{capacity} of the goal space should be quite low, and hence not much information should propagate through to the LLC.
To test this hypothesis, we implemented two variants of H2O2 with unconditional LLCs.

The first one is "Autopilot (BC)", where only the Behavioral Cloning policy, $ \mu(a_t | b_t) $ (notice how the dependence of the goal $g$ is dropped), is used for acting and trained by maximising the log-likelihood of actions. These action come from trajectories drawn from the replay and are generated while the full agent is acting. We note that there is no special goal selection or external reward used to train the "Autopilot (BC)". This LLC variant is simply learning to re-generate experience it has already seen, and the HLC is only requred to select when to deploy the LLC, rather than selecting which goals it should attend.

The second variant is "Autopilot (Vtrace)", and is trained with the same offline RL method described in \cref{sec:llc} (\textbf{Offline RL} paragraph), but without the conditioning on the goal (see \cref{eq:llc-vtrace}). This means that the policy of the LLC is only conditioned on the current recurrent state, $b_t$, as all dependencies on a goal, $g$, are dropped. The data is drawn similar to "Autopilot (BC)", but now the Advantage is calculated based on the external reward. For this reason, "Autopilot (Vtrace)" is actually trained using offline data to solve the task directly, and the HLC simply chooses when to deploy the LLC.

\begin{figure}[htb!]
    \centering
    \includegraphics[width=0.8\columnwidth]{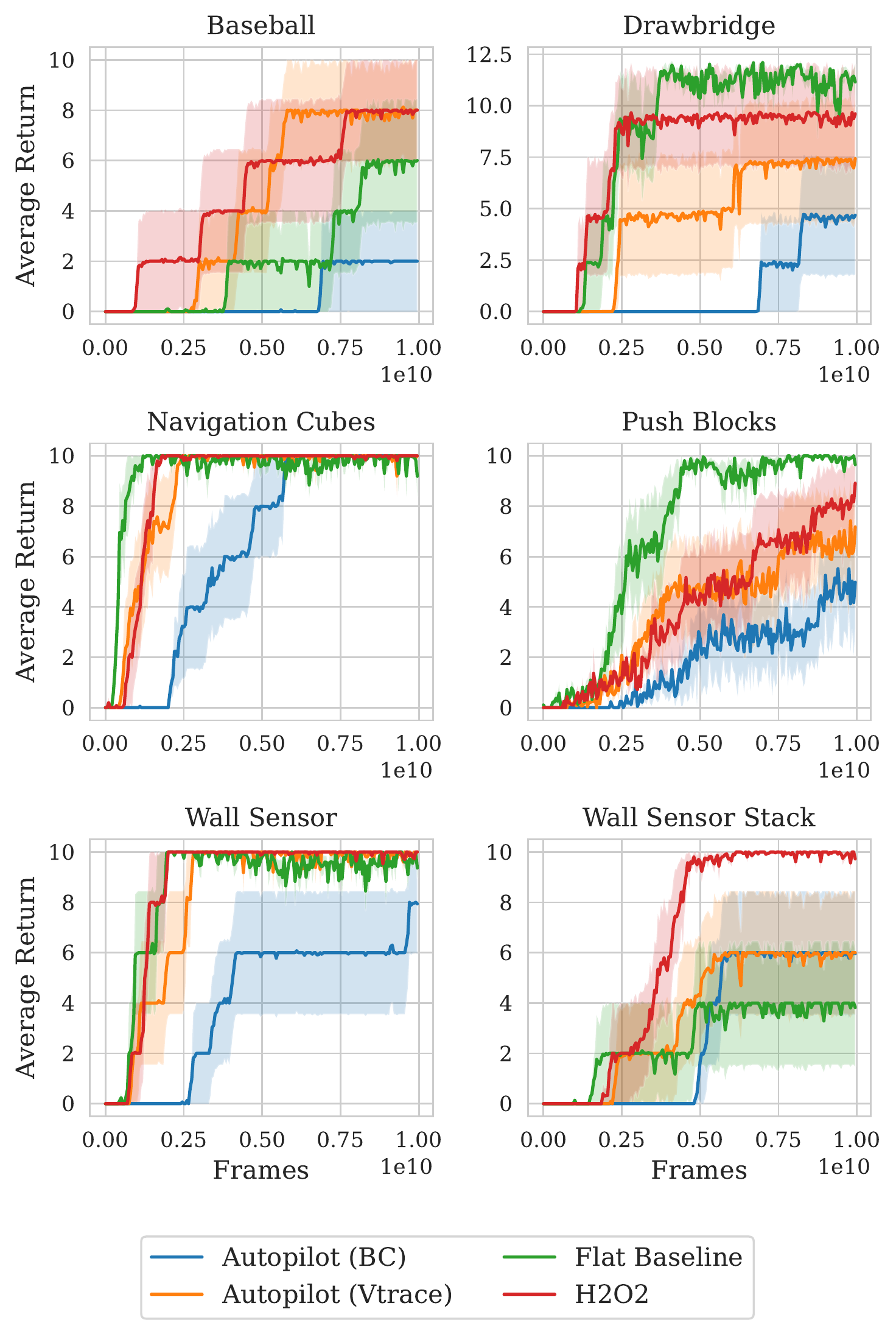}
    \caption{Average episode return per task for different agents. We only show performance on the 6 tasks that agents made some progress, and exclude the results on \texttt{Throw Across} and \texttt{Remember Sensor}, where no agent made any progress on.}
    \label{fig:mean_return_per_episode_main_all_variants}
\end{figure}

We compare their performance on a per-task basis with the best performing H2O2 variant used in the main report and the Flat baseline in \cref{fig:mean_return_per_episode_main_all_variants}. As we can see, both of these variants with unconditional, reward-maximising LLCs are performing worse than H2O2 on 3 of the tasks, and about as well on the other 3. In addition, "Autopilot (Vtrace)" performs better than the "Autopilot (BC)" on 4 out of 6 tasks, with the latter matching performance on the remaining two. This is not very surprising as the V-trace loss allows us to consume sub-optimal, off-policy data, whereas Behavioral Cloning is expected to work best when using predominantly expert data.

What is somewhat surprising though, is that there is a significant difference between the H2O2 variant with a strongly regularized goal space and an unconditional LLC. This means that even though the goal space is strongly regularized, it is still able to capture enough information and propagate it to the LLC policy. An additional final reason of the much lower performance is becayse none of the two Autopilot variants make use of any of the auxiliary losses  mentioned in \cref{sec:llc}, which could mean that they play a significant role in shaping the LLC network.

\begin{figure}
\includegraphics[width=0.8\columnwidth]{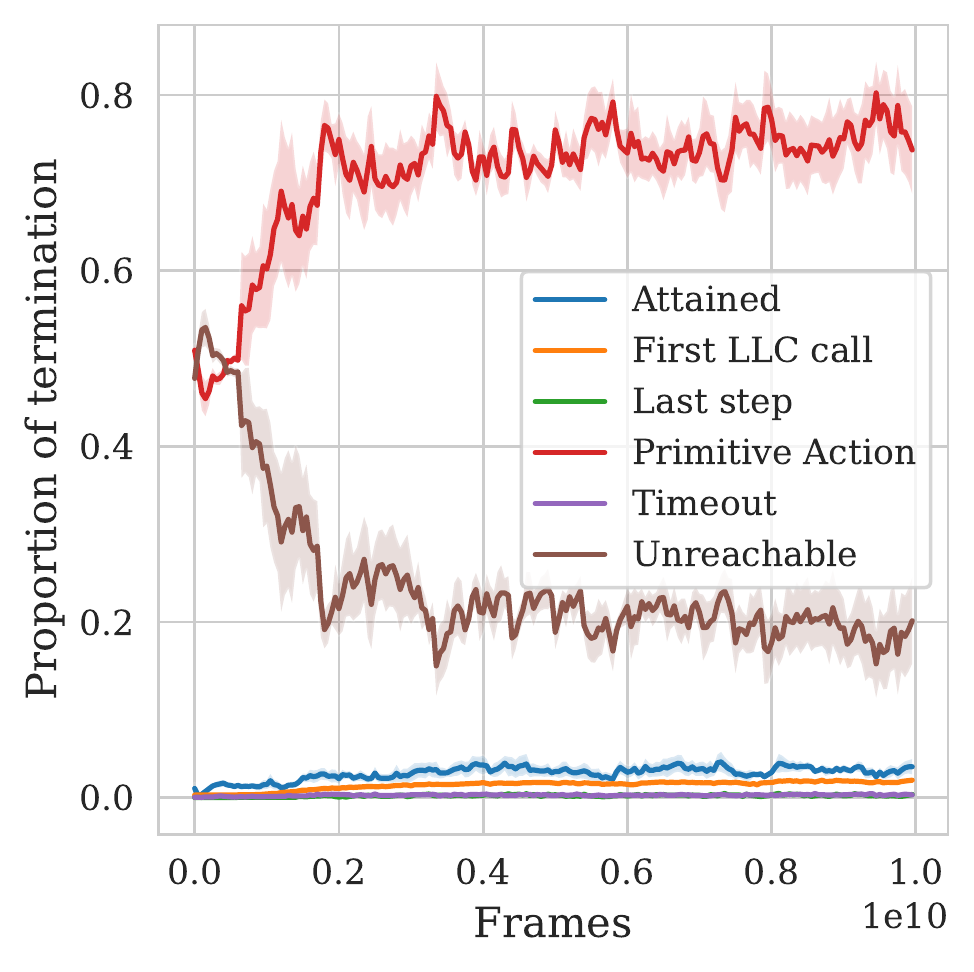}
\caption{Proportion of LLC execution termination reasons as training progresses.}
\label{fig:max_llc_steps_32_termination_reason}
\end{figure}

\subsection{Compare the number of model parameters}
\label{sec:experiments:more_parameters}
We cannot attribute the increased performance solely to more parameters.
Increasing the number of parameters of the flat baseline increases performance in some of the hard eight tasks, but not every way to increase parameters is effective.
Even a flat baseline with more parameters fails to outperform H2O2 in one of the tasks.

\begin{figure}[htb!]
    \centering
    \includegraphics[width=0.9\columnwidth]{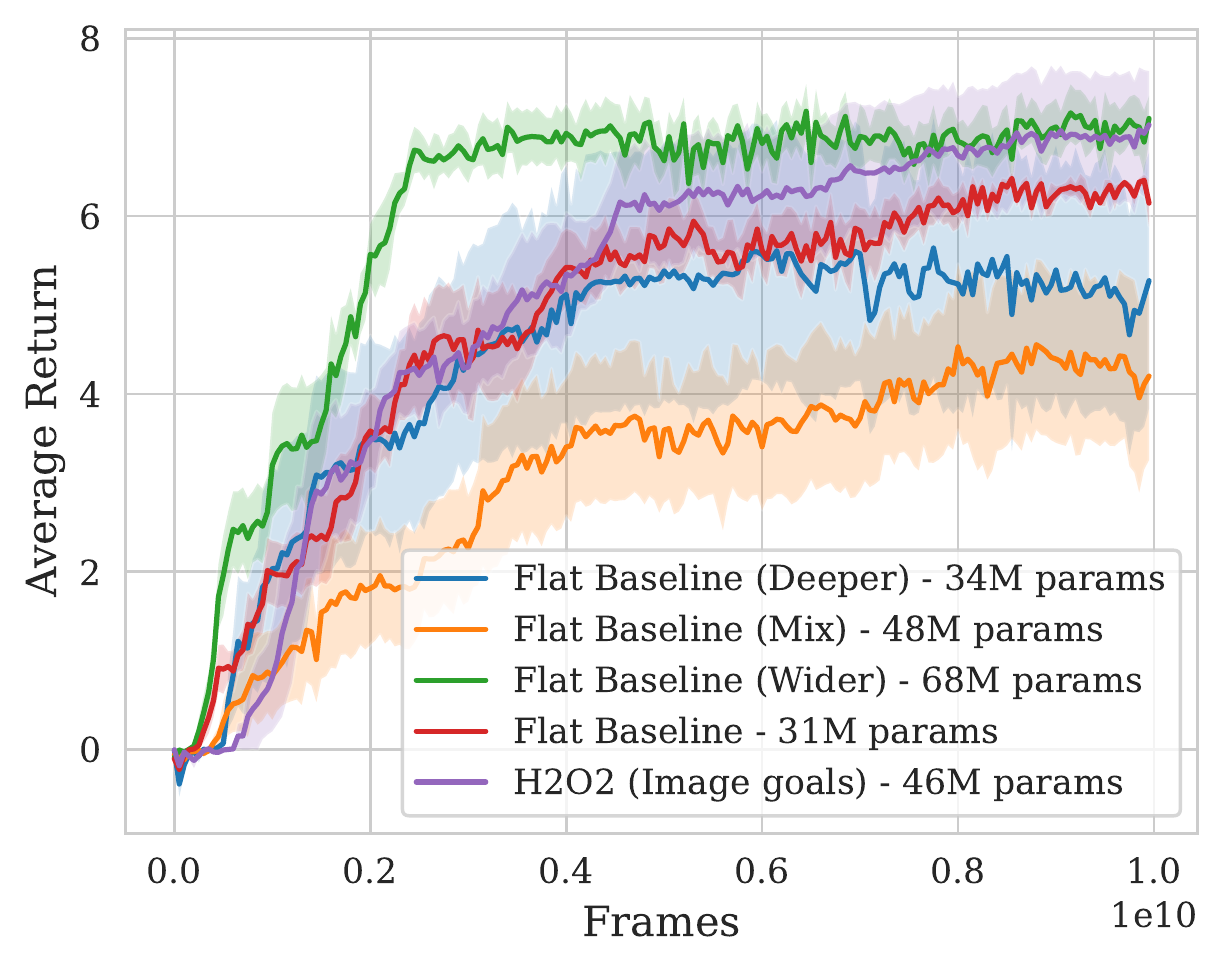}
    \caption{Average performance of H2O2 and flat baselines with different number of parameters.}
    \label{fig:flat_baseline_size_ablation}
\end{figure}

We compared the performance of H2O2 against flat Muesli baselines with different number of parameters\footnote{All numbers of parameters reported are rounded to millions.} in \cref{fig:flat_baseline_size_ablation}.
The H2O2 has 46M parameters (34M on the HLC, 12M on the LLC).
The flat baseline presented in the paper so far has 32M parameters.
We additionally experimented with a baseline with additional MLP layers (``Deeper'', 35M parameters), and a baseline with wider hidden layers (``Wider'', 68M parameters) and a baseline with similar number of parameters (``Mix'', 48M parameters), achieved by both adding more layers and making the hidden layers wider.
\Cref{fig:flat_baseline_size_ablation} shows the performance of the different agents across the six tasks where performance was nontrivial, throughout training.
The ``Deeper'' flat baseline is an example where we increasing parameters but achieve worse performance.
The ``Wider'' flat baseline has more parameters than H2O2 and matches or outperforms the flat baseline across the board.
H2O2 outperforms the flat baseline in \texttt{Baseball} and \texttt{Wall Sensor Stack}.
The ``Wider'' baseline outperforms H2O2 in \texttt{Baseball}, but is outperformed in \texttt{Wall Sensor Stack}.
Surprisingly, the ``Mix'' baseline performed worst than any other agent, even though it has the same number of parameters as H2O2.

We think that these results go to show that the number of parameters is not the whole story, and that carefully tuning the architecture hyper-parameters, such us number of layers and their size, can have a far greater impact. 

\subsection{Comparing against Action Repeat}
\label{sec:action_repeat}

\begin{figure}[htb!]
    \centering
    \includegraphics[width=0.8\columnwidth]{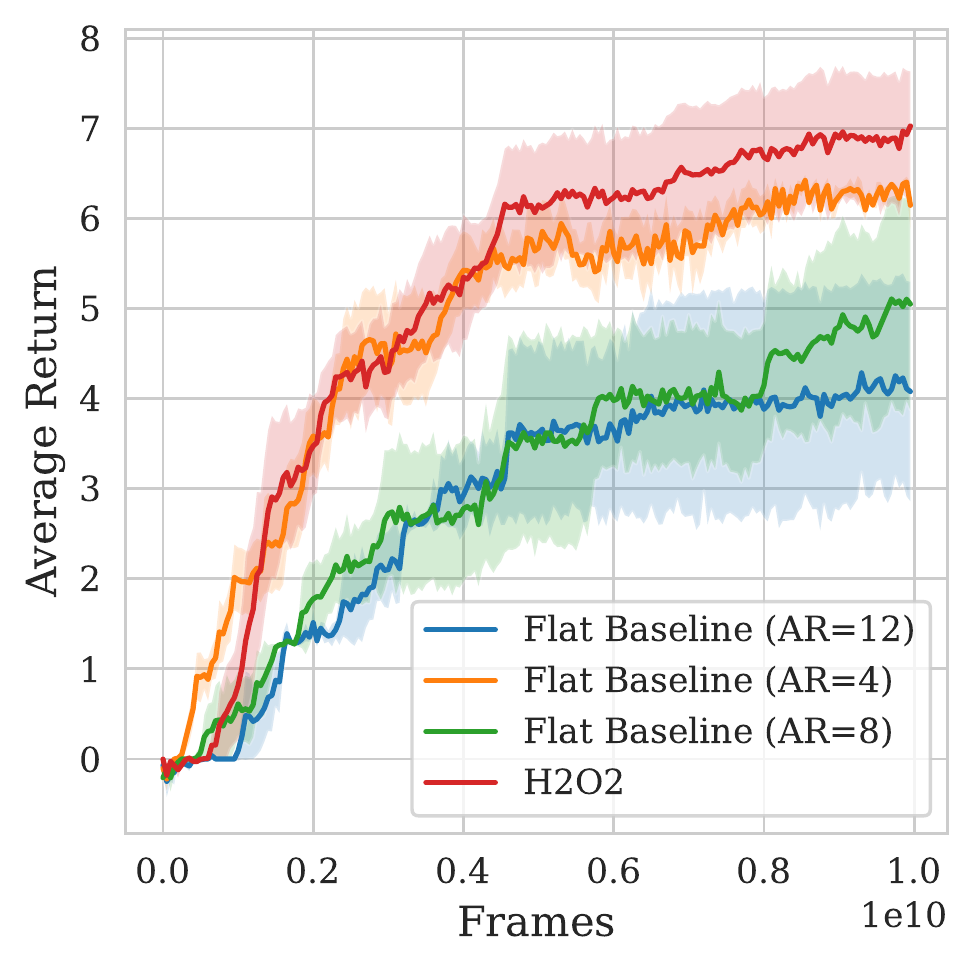}
    \caption{The average return across levels for different action repeats (AR).}
    \label{fig:mean_return_env_steps_ablation}
\end{figure}

One could claim that the reason for H2O2's strong performance is that it uses options lasting about $2$ timesteps, which enable the HLC to utilize an effective action repeat of $8$, rather than the actual action repeat of $4$ used.
To examine this claim, we have run experiments with flat baselines that instead use action repeats of $8$ and $12$. 

As we can see in \cref{fig:mean_return_env_steps_ablation} though, the claim above is not supported by evidence, since the flat baselines with larger action repeats perform significantly worse.
A possible reason for this is that conventional action repeats simply repeat the same action for the specified number of steps, which means that the agent loses the ability to execute different actions over that duration.
This can be catastrophic for many tasks that require finer motor control, and by doing so these agent loose any benefits they might gain in terms of shorter value backups, when learning their value function. 
On the contrary, since short options seem to perform at a SOTA level, it means that the primitive actions executed within the options are not simple action repeats, but most likely combinations of different primitive actions, depending on the current state of the agent. 

\subsection{Agent videos}
\label{sec:agent-videos}

We generated videos
\footnote{\url{https://youtube.com/playlist?list=PLlHafZmkZCGmIYSHO9aot75l07kbiuLN0}}
of a trained H2O2 agent solving the different tasks from the Hard Eight suite of tasks. The video shows multiple episodes from the agent's first-person view.

The frames in red correspond to the steps where the HLC selects an SMDP action (a primitive action or a goal), whereas the other frames refer to MDP steps where actions are selected by the LLC in response to the HLC's SMDP action.
When the LLC is executing an option per se (i.e.~following a goal), the latent goal selected by the HLC is shown in the video in the ``Latent Goal'' box.
The latent goal is a vector in $[-1, 1]^{32}$ and is visualized as a bar plot.

The ``Option State Value'' box shows the value estimate $\widehat{V}^{\llc}$ for the given goal, over time.
The rightmost point in the plot corresponds to the value estimate for the most recent timestep in the video, and the flat red line is the threshold for unattainable goals (note that points below the line are a plotting artifact as they would refer to timesteps before the option started).
This plot is only valid when an option is being followed (that is, when a Latent Goal is shown).

The ``Distance-to-Goal Predictor'' box shows the probability estimates of the predictor for how many steps into the future the goal is attained.
This plot is only valid when an option is being followed (that is, when a Latent Goal is shown).

We can see in the videos that both the Option State Value and the Distance-to-Goal Predictor probabilities behave sensibly.
The option state value rises as the agent approaches the goal, and the mass over distance-to-goal shifts towards the zero bin.
Once the probability mass is maximum at zero, the option terminates, as specified by the termination criterion.

The majority of HLC actions in the videos are primitive actions, and when goals are issued they typically last for a few steps (three to seven).
The goal-following behavior we observed in the videos seem to cover navigating towards objects, picking them up and carrying them, whereas primitive actions seem to be reorienting the agent's view.

Qualitatively, we see from the video different signs that the hierarchical behavior being learned is sound and in line with what we would expect to be useful for solving the tasks, but that specific parts of the agent's behavior are still executed in terms of primitive actions rather than options.

\subsection{Low Level Controller Heatmaps}

\begin{figure*}
  \centering
  \includegraphics[width=0.9\textwidth]{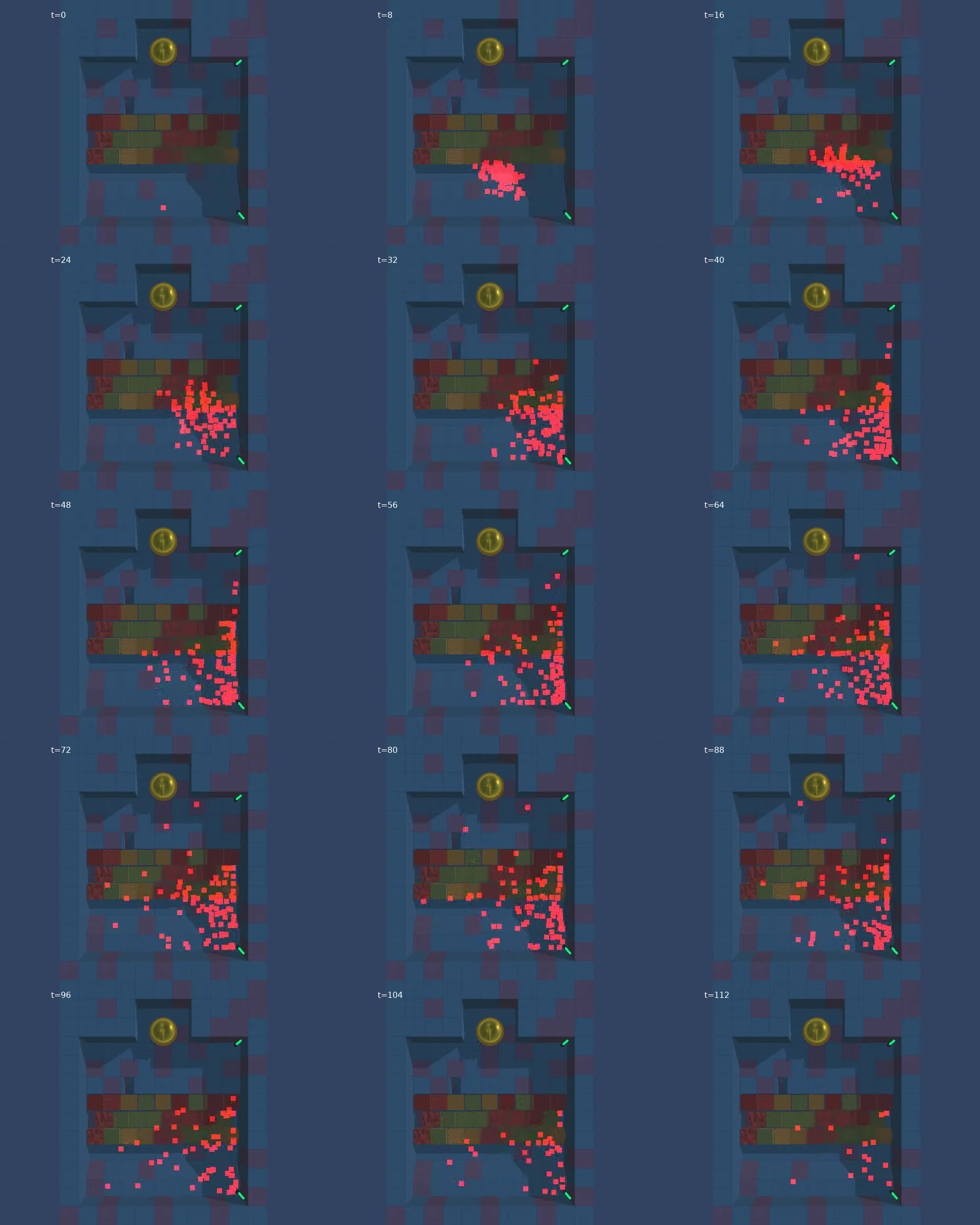}
  \caption{Top-down positions of the LLC at different timesteps in the \texttt{Navigation Cubes} task. Note that there is no HLC here, as we are manually giving the same image goal of a golden apple to the LLC, for 20 repetitions. Note how all LLC instances are following a similar path towards the golden apple, with some variation. Each red dot corresponds to the position of a single LLC instance out of 100 identical LLCs initialized from the same state and given the same goal. Note that when each LLC terminates for all 20 repeated goals, or the episode terminates by reaching the apple, its corresponding red dot disappears.}
  \label{fig:heatmap}
\end{figure*}

In \cref{fig:heatmap}, we present a qualitative analyse of the goal-following performance of the LLC, in isolation from the HLC. The setup is the following: 100 instances of the LLC are initialized in different instances of the same evaluation \texttt{Navigation Cubes} task. Each of them is shown a goal image of a golden apple being right in front of the agent, which would have been the observation had the agent reached the golden apple. The LLCs then take actions, trying to reach the given goal. If the LLC option execution is terminated, the same goal image is presented again, for a total of 20 times per episode, for each LLC instance. We record the trajectory of every LLC instance and overlay them all on the same top-down map of the environment, where each LLC instance can be seen as a single red dot. \Cref{fig:heatmap} shows snapshots of the overlaid trajectories at different timesteps.

At timestep 0, all instances start from the same position, and initially they move in a similar direction. After about 40 timesteps, their trajectories start to differentiate significantly, with some of them making it across the cube pit, while others end up spending a long time around the corner of the room. The main takeaway from this analysis is that indeed the different LLC instances behave similar in the beginning, but their trajectories seem to diverge at longer timescales, as they interact with the environment.

\end{document}